\documentclass{article}

  \PassOptionsToPackage{numbers, compress}{natbib}

    \usepackage[preprint]{neurips_2019}

\usepackage[utf8]{inputenc} 
\usepackage[T1]{fontenc}    
\usepackage{hyperref}       
\usepackage{url}            
\usepackage{booktabs}       
\usepackage{amsfonts}       
\usepackage{nicefrac}       
\usepackage{microtype}      
\usepackage{graphicx}
\usepackage{amsmath}
\usepackage{algorithmic}
\usepackage[]{algorithm2e}

\title{Inherent Noise in Gradient Based Methods}

\author{%
  Arushi Gupta \\
  \texttt{arushig@princeton.edu} \\
}

\begin{document}

\maketitle

\begin{abstract}
Previous work has examined the ability of larger capacity neural networks to generalize better than smaller ones, even without explicit regularizers, by analyzing gradient based algorithms such as GD and SGD. The presence of noise and its effect on robustness to parameter perturbations has been linked to generalization. We examine a property of GD and SGD, namely that instead of iterating through all scalar weights in the network and updating them one by one, GD (and SGD) updates all the parameters at the same time. As a result, each parameter $w^i$ calculates its partial derivative at the stale parameter $\mathbf{w_t}$, but then suffers loss $\hat{L}(\mathbf{w_{t+1}})$. We show that this causes noise to be introduced into the optimization. We find that this noise penalizes models that are sensitive to perturbations in the weights. We find that penalties are most pronounced for batches that are currently being used to update, and are higher for larger models.\end{abstract}

\section{Introduction}

Previous work has shown that neural networks with large capacity, even in the absence of explicit regularization, generalize better than smaller capacity networks. 
\citet{neyshabur2014search} suggested through analogy to matrix factorization that network size is not the main form of capacity control in neural networks.  
\citet{zhang2016understanding}  then demonstrated that neural networks are capable of memorizing random labels, but still generalize given good data. These findings prompted investigation into stochastic gradient descent's ability to implement some form of regularization that allows larger architectures to outperform smaller ones, even in the absence of explicit regularization, such as dropout, batch normalization, and weight decay  \citet{srivastava2014dropout}\citet{ioffe2015batch}\citet{krogh1992simple}.

One line of inquiry has studied how noise may improve generalization ability.  \citet{an1996effects} studied the effect of adding noise to backpropagation. 
\citet{blundell2015weight} found that training so that the weights learn to cope with uncertainty improves generalization. Later, \citet{mandt2016variational} noted that when training with SGD, each minibatch of size $S$ provides $S$ independent samples of the gradient.  Letting $w_t$ be the weights at time $t$, $\hat{L}$ the training loss, and $\eta$ the learning rate, \citet{mandt2016variational} describe the SGD update as
 \begin{equation}
w_{t+1} = w_t  - \eta \nabla \hat{L}(w_t) + \eta \epsilon_t
\end{equation}
where $\epsilon_t$ has zero mean and some covariance, and is referred to as the noise induced by  minibatch sampling. It was later discovered that the noise in SGD is anisotropic, yielding study of the gradient noise when the covariance matrix of $\epsilon_t$ is not constant \citet{zhu2018anisotropic}.

Related to the idea of noise improving generalization performance are parameter perturbations. Parameter perturbations are a tool used in PAC Bayes bounds \citet{dziugaitecomputing}, which include a term measuring the 'sharpness' of the final minimum found by training. The 'flatness' of the minima of the training loss relates to the volume of the space around the final minimizer that has a loss similar to the actual minimizer. \citet{keskar2016large} found that flat minimizers tend to be more robust to noise introduced by parameter perturbations, and that large batch training produces sharper minimizers than small batch training.  Noise has also been used as an explanation for explicit regularization such as dropout \citet{wager2013dropout}.

In this paper, we consider how the inherent noisiness of using a gradient based optimizer along with capacity may contribute to generalization for neural networks. In particular, we notice that instead of iterating through each scalar parameter and updating them one by one, GD updates all the parameters at the same time. As a result, parameter $w^i_t$ calculates its partial derivative at the stale parameter vector $\mathbf{w}_t$, but then suffers loss $\hat{L}(\mathbf{w_{t+1}})$.

We find a term to describe the above noise, and find that the optimization introduces a penalty for solutions that are sensitive to parameter pertubation. We then relate it to the Taylor series of the loss and compare the first order approximation to the loss made by SGD to the actual change in loss.  We find that for larger models, although they may overfit more in a first order sense, this implicit penalty is also higher potentially producing a regularization effect.

\subsection{Related Work}

There has been a line of inquiry about the dot product of the gradients during SGD training.   \citet{sankararaman2019impact}  noted how width and depth affect a quantity they call 'gradient confusion,' and determine how this affects the speed of convergence of SGD. Others \citet{arpit2017closer} have measured the 'loss' sensitivity for different capacity networks for good versus corrupted data.
Several works have examined whether neural networks learn 'simpler' functions before learning more complex ones \citet{kalimeris2019sgd} \citet{rahaman2018spectral}.

In order to study the implicit regularization provided by SGD, one line of work has examined the 'flatness' or 'sharpness' of the minima found by SGD \citet{hoffer2017train}, with the hypothesis that flatter minima generalize better. Other work has posited that the ratio of learning rate over batch size is important in SGD optimization \citet{jastrzkebski2017three}. Other work has analyzed the anisotropic nature of the noise in SGD \citet{zhu2018anisotropic}. \citet{dinh2017sharp} examine whether sharp minima for neural networks can generalize, and conclude that flatness must be defined carefully. \citet{dziugaitecomputing} took a PAC Bayes approach to computing generalization bounds. \citet{neyshabur2018role} studied the effect of over-parameterization on generalization by looking at 'unit capacity' and 'unit impact' for 2 layer ReLU networks. Other work has empirically studied how network width may affect the 'noise scale' of the network \citet{park2019effect}.

Other work has examined the local elasticity of neural networks \citet{he2019local}, that is the ability of one data point to alter the prediction on another.
\citet{novak2018sensitivity} has investigated the input output Jacobian and concluded that neural networks are more robust in the data manifold.

  \subsection{Preliminaries}
  
  We use $[d] = \{1,2,3,...,d\}$. We denote by $L(w_t)$ the true (general) loss associated with the weights of the neural network, $w_t$ at a certain time t. That is,  let $(x,y)$ be training data points and labels, such that $x \in \mathcal{X}$ and $y \in \mathcal{Y}$ be data drawn from some distribution, $D$, and let $\ell(x, y, w_t)$ be some loss function, then $ L(w_t) = \mathbb{E}_{(x , y)\sim D} [\ell(x, y, w_t)]$. We will sometimes omit the $y$ term and write $\ell(x, w_t)$, where the $y$ corresponding to the $x$ is taken implicitly. 
   We  describe the training loss, which is the average loss over the training data as
   \begin{equation}
   \hat{L}( \mathcal{T}, w_t) = \frac{1}{|\mathcal{T}|} \sum_{x_i \in \mathcal{T}} \ell( x_i, w_t)
   \end{equation}
   where $\mathcal{T}$ is a set containing the training data points.

 \textbf{SGD:} We take $\hat{L}(B_i, w)$ to be the empirical loss evaluated on the $i$th minibatch, $B_i$. 
Instead of taking the full gradient update over $\mathcal{T}$, SGD computes 
 \begin{equation}
 w_{t+1 } = w_t - \eta \nabla \hat{L}(B_i, w_t )
 \end{equation}
 Typically, the learning rate schedule $\eta$ is manipulated, however, for our experiments and analysis we maintain a fixed constant learning rate, so that we may separate the effects of the learning rate schedule from the effects of batching and SGD. Although SGD is explicitly given (and told to minimize) the training loss, without direct knowledge of the true loss, in practice it often manages to find a solution that has reasonable generalization loss.

 \section{Our model}
 
 \textbf{Simultaneous move games} GD updates all the scalar weights at the same time instead of updating them individually. Each scalar weight therefore knows the values of the other weights at $\mathbf{w_t}$ but is then evaluated at $\mathbf{w_{t+1}}$. An analogy to this process is the game of synchronous chess, where $w^i$ and $w^j$ are players, and each player must make their move based on the current state of the board, but simultaneously without knowledge of the other player's concurrent move (of course, in synchronous chess each player may try to 'guess' what the other player will do, whereas the weights do not).
 \subsection{Parameter updates at the same time}
Because we are considering GD in this section, we sometimes omit the data parameter of the loss, since it is always $\mathcal{T}$. In contrast to gradient descent, consider the following algorithm:

\begin{algorithmic}
\FOR{t $\in \{1,2,...\}$}
\FOR { $w^i_t$ in $ w_t$}
\STATE $w^i_{t+1} = w^i_t - \frac{ \partial \hat{L}}{\partial w^i}(\mathcal{T}, w^1_{t+1},w^2_{t+1}...,w^i_t, ...,w^d_t)$
\ENDFOR
\ENDFOR
\end{algorithmic}

In other words, this algorithm takes the partial derivative of each scalar weight, and updates one of them at a time, instead of updating them all at the same time. The $w^i$ are optimized jointly, so that each knows what the current values of the others are when it makes its decision on how to update. 
The change in loss experienced by weight $w^i$ is $ \hat{L}( w^1_{t+1},w^2_{t+1}...,w^i_t, ...,w^d_t)  - \hat{L}( w^1_{t+1},w^2_{t+1}...,w^i_{t+1}, ...,w^d_t) $. Notice that the only weight that changes is $w^i$, so  when $w^i$ updates itself there is no uncertainty introduced by the other weights $w^j$. 

Gradient descent, by contrast, computes all the gradients at the old weights, $\mathbf{w_t}$ as follows:
\begin{algorithmic}

\FOR{t $\in \{1,2,...\}$}
\FOR { $w^i_t$ in $ w_t$}
\STATE $w^i_{t+1} = w^i_t - \frac{ \partial \hat{L}}{\partial w_i}(\mathcal{T}, w^1_{t},w^2_{t}...,w^i_t, ...,w^d_t)$

\ENDFOR
\ENDFOR

\end{algorithmic}
However, following these updates, each weight suffers a loss 
\begin{equation}
\hat{L}(w^1_t,...,w^i_{t},...,w^d_t) - \hat{L}(w^1_{t+1},...,w^i_{t+1},...,w^d_{t+1})
\end{equation}
Each weight $w^i$ computed its partial derivative at $\mathbf{w_t}$, and therefore had full information about the other parameters at time $t$. However, because all the parameters are combined to produce a single model with $\mathbf{w_{t+1}}$, an implicit penalty is introduced for weight changes $w^i_t \rightarrow w^i_{t+1}$, that were not robust to perturbations made by the other weights. More specifically, if all weights updated in the same round, and no uncertainty were introduced by any of the weights $w^j$ for any updating weight $w^i$, the change in loss at the end of the round would be:
\begin{equation}
 \sum_{i \in [d]} \hat{L}(w^1_t,...,w^i_{t},...,w^d_t) - \sum_{i \in [d]} \hat{L}(w^1_t,...,w^i_{t+1},...,w^d_t) 
\label{eq:realpen}
\end{equation}

but in actuality, 
GD first combines the various weight updates into a single model with weights $w^1_{t+1},...,w^d_{t+1}$, and then produces a joint penalty as follows:

\begin{equation}
\begin{split}
&  \underbrace{ \sum_{i \in [d]} \hat{L}(\mathbf{w_t}) - \sum_{i\in [d]} \hat{L}(w^1_t,...,w^i_{t+1},...,w^d_t) }_{\text{ first term}}  \\
&   + \underbrace{\left[  \hat{L}(\mathbf{w_t}) - \hat{L}(w^1_{t+1},... w^i_{t+1},...,w^d_{t+1})  -  \left(\sum_{i \in [d]} \hat{L}(\mathbf{w_t}) -  \sum_{i \in [d]} \hat{L}(w^1_t,...,w^i_{t+1},...,w^d_t)   \right) \right]}_{\text{second term}}
\end{split}
\end{equation}

The first term is the objective function, and searches for weights $w^i_{t+1}$ that would most improve the loss if no uncertainty were introduced by any weight $w^j$ for any other weight $w^i$.
The second term can be thought of as a regularizer, or penalty. It will reward weight choices $w^i_t \rightarrow w^i_{t+1}$ whose effect on the loss is similar or better when they are implemented alongside other parameter updates than when they are implemented individually.
These effects apply to the discrete dynamics of GD. Namely, if the learning rate is small enough, it may be close to the case that the other parameters don't change very much. 

Notice that if the loss were to behave linearly over this round:
\begin{equation}
 \begin{split}
     &=  \left(\hat{L}(w^1_{t} - w^1_{t+1},... w^i_{t} - w^i_{t+1},...,w^d_{t} -w^d_{t+1})     - \sum_{i \in [d]} \hat{L}(0,...,( w^i_{t} - w^i_{t+1}) ,...,0)    \right) \\
    & = 0
  \end{split}
\end{equation}

So that linear models, where no uncertainty is introduced by any weight $w^j$ for $w^i$, would not receive a penalty. We will be interested in experimentally examining the effect of this penalty for SGD. To do so, we will create a Taylor approximation to the loss and measure the first order effects versus the higher order effects, but first we discuss why the above penalty may link to generalization. 

We notice that larger models have more nodes, and hence have a propensity to behave more non-linearly, and a potential ability to claim higher rewards from Equation \ref{eq:realpen} without generalizing well. However, we hypothesize that any undesirable non-linear behavior will be curbed by producing a higher value of the penalty above. We reason that if large models are regularized more using this mechanism, they may achieve better generalization performance.

\subsection{Penalizing functions not robust to perturbation}
\label{sec:pen}

There is a rich set of literature relating noise to generalization. Consider, for a counterexample, a decision tree, which is prone to overfitting unless ensembled. From Elements of Statistical Learning \citet{friedman2001elements} p. 307
for splitting variable $j$ and split point $s$, the split point can be decided according to the following optimization problem:

\begin{equation}
\min_{j,s} \left[ \min_{c_1} \sum_{x \in R_1(j,s)} (y_i - c_1)^2 + \min_{c_2}  \sum_{x_i \in R_2(j,s) } (y_i - c_2)^2\right]
\end{equation}

This optimization problem gives the tree a greedy, but precise look at the loss \emph{after} the update, and it may choose the $j,s$ that produce the best value of the loss \emph{a posteriori}. The optimization is therefore not inherently noisy, since $j$ is aware of exactly which $s$ it will be paired with and has access to the resulting loss, and the penalty term described in the previous section does not apply. The neural network, by contrast, cannot for example try all possible weight vectors $\mathbf{w_{t+1}}$ such that $||\mathbf{w_{t+1}} - \mathbf{w_t}||<\eta$ and select the one that produces the lowest loss. 

In particular, due to the partial derivative, $w^i$ expects the weight vector to move from $\mathbf{w_t}$ to $(w^1_t,...w^i_{t+1},...w^d_t)$, but in reality it moves from $\mathbf{w_t}$ to $\mathbf{w_{t+1}}$. The movement of the other parameters can be seen as a perturbation to the update made by $w^i$.
Therefore, from the perspective of $w^i$, its loss at time $t+1$ is:

\begin{equation}
\hat{L}(w^i_{t+1}) = \hat{L}(w^i_{t+1} | \mathbf{w_t}) + \epsilon_t
\end{equation}

where $\epsilon_t$ models the effect on the loss due to other weights changing and $\hat{L}(w^i_{t+1}| \mathbf{w_t})$ is the value of the loss when the optimizer chooses $w^i_{t+1}$ assuming all other weights remain at their time $t$ values. We would expect that a larger $\eta$ would produce a larger perturbation, and could increase the magnitude of $\epsilon_t$. Although larger models tend to have closer distance to initialization, so that $||\mathbf{w_{t+1}} - \mathbf{w_t}||$ could be smaller, larger models have more weights and more possible activation patterns, which could still cause the loss perturbation $\epsilon_t$ to be large. Unlike a Gaussian perturbation, $\epsilon_t$ is driven by the data, so it is not unreasonable to expect $w^i$ to be able to withstand it.

Penalizing weight changes that were not robust to other parameters in the network being simultaneously perturbed could qualitatively bias the network towards flatter minima, which reflect weight settings which are not too sensitive to perturbation.

\subsection{Expected behavior on experiments}
\label{sec:expbev}
We run our experiments with SGD, not GD, so that we may observe the interaction of the penalty with the stochasticity introduced by SGD. Notice that the penalty described can be taken on a particular batch. If a batch $B_u$ is used to update $w_t \rightarrow w_{t+1}$, we would expect each weight $w^i$ to successfully make progress on $B_u$  if only $w^i$ were to update. Therefore, $\sum \hat{L}(B_u, w^1_t,...,w^d_t) - \sum \hat{L}(B_u, w^1_t,...w^i_{t+1},...w^d_t) $ would be high.  However, we would also expect that because $w^i_t$ has access to the the other weights $w^j_t$ along with the particular activations produced by the data $B_u$, that the weight change $w^i_t \rightarrow w^i_{t+1}$ may have a larger penalty on $B_u$ than on other batches.
For a batch, $B_a$, that updated long ago, we would expect $\sum \hat{L}(B_u, w^1_t,...,w^d_t) - \sum \hat{L}(B_u, w^1_t,...w^i_{t+1},...w^d_t) > \sum \hat{L}(B_a, w^1_t,...,w^d_t) - \sum \hat{L}(B_a, w^1_t,...w^i_{t+1},...w^d_t) $, but we would also expect the penalty on $B_a$ to be smaller, as $\mathbf{w_t}$ is less likely to be very overfitted on a batch that updated long ago. We expect that recently updating batches, $B_r$, may display an intermediate behavior.

   \begin{figure*}[ht]
\centering
\includegraphics[scale=0.26]{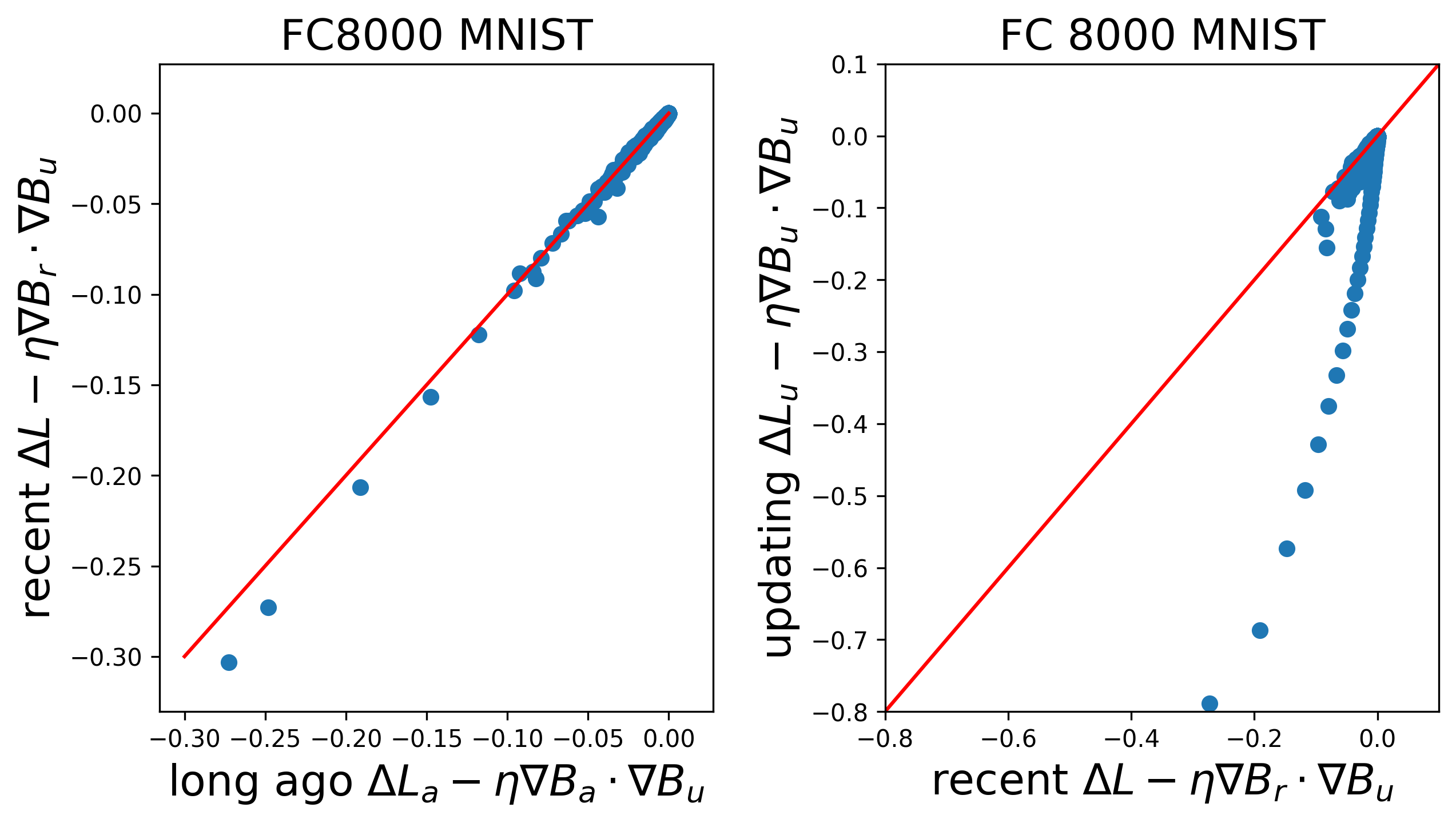}
\includegraphics[scale=0.26]{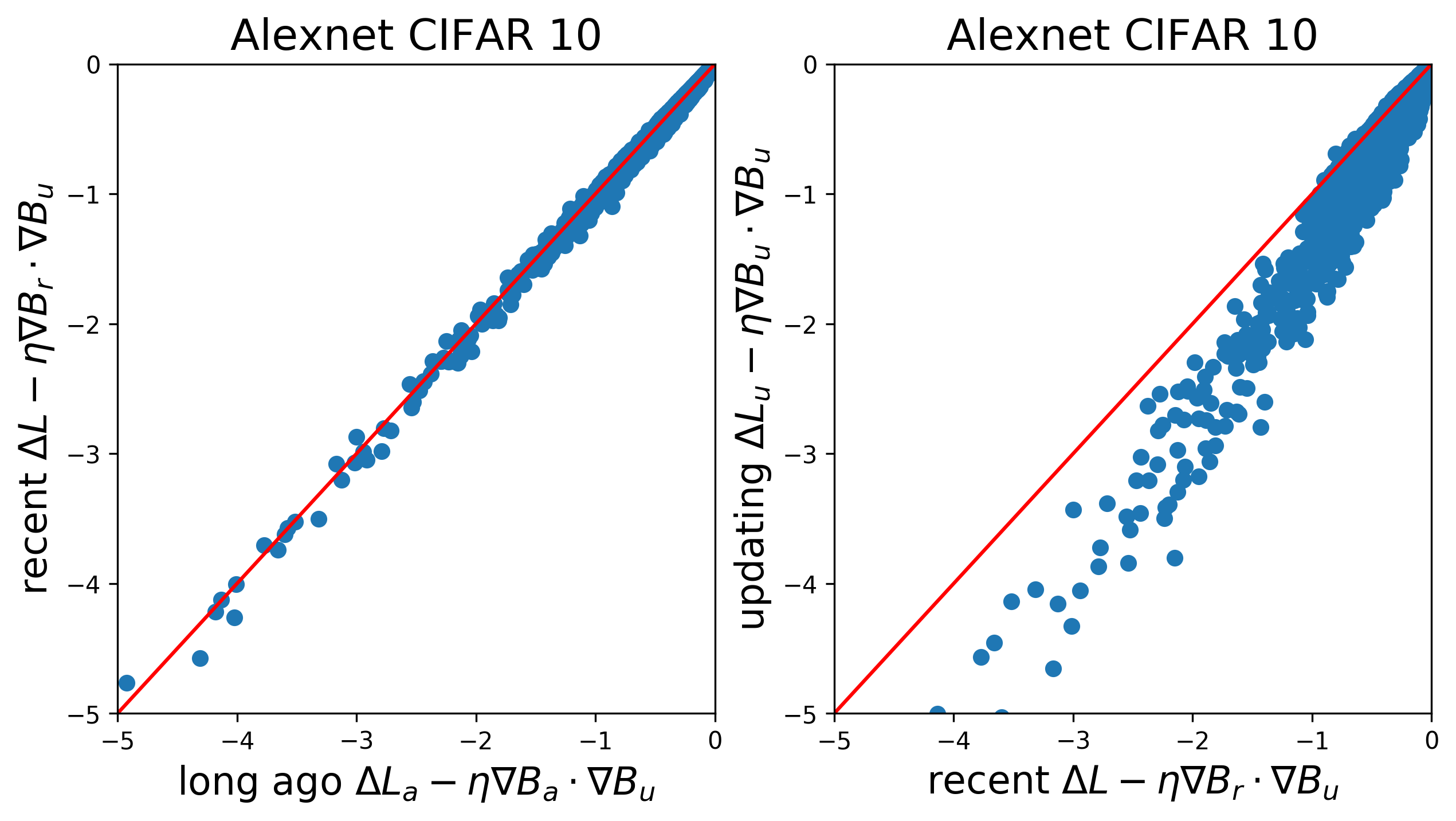}
\includegraphics[scale=0.26]{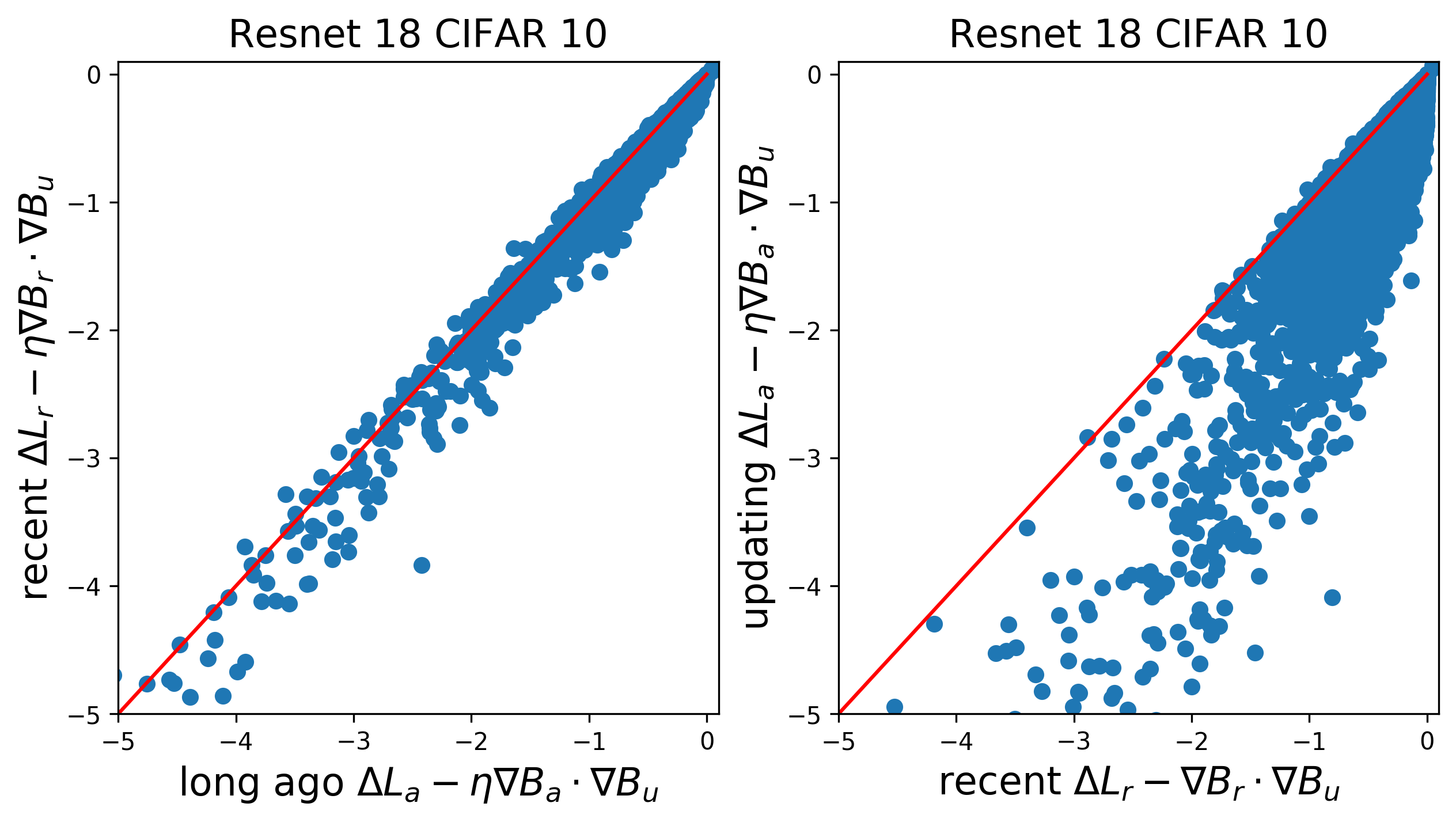}
\caption{ Red line depicts y=x.  \textbf{Row 1}(from left to right): $-\hat{L}^h_{B_r}$ versus $-\hat{L}^h_{B_a}$for FC 8000 on MNIST followed by $-\hat{L}^h_{B_u}$ versus $-\hat{L}^h_{B_r}$ for FC8000 on MNIST  followed by $-\hat{L}^h_{B_r}$ versus $-\hat{L}^h_{B_a}$ for Alexnet on CIFAR 10 followed by $-\hat{L}^h_{B_u}$ versus $-\hat{L}^h_{B_r}$ for Alexnet on CIFAR 10. \textbf{Row 2 }$-\hat{L}^h_{B_r}$ versus $-\hat{L}^h_{B_a}$ for Resnet 18 on CIFAR 10 followed by  $-\hat{L}^h_{B_u}$ versus $-\hat{L}^h_{B_r}$ for Resnet 18 on CIFAR 10. We find that penalty is more negative for updating batch, $B_u$. }
\label{fig:nondot}
\end{figure*}

  \subsection{Taylor series on the weights}

 We will use the Taylor series so that we find a way to experimentally measure the penalty. We will be interested in the behavior of different batches, as well as different capacity models. We will use $\nabla \hat{L}(B_i,w)$ and $\nabla B_i$ interchangeably.  
 We consider the effect that moving $- \eta \nabla B_1$ has on the loss of another batch, $B_2$:
 \begin{equation}
 \begin{split}
 \hat{L}(B_2, w - \eta \nabla B_1) \approx \hat{L}(B_2, w) - \eta \nabla \hat{L}(B_2, w)\cdot \nabla B_1(w) + \hat{L}^h_{B_2}(\nabla B_1)
 \end{split}
 \label{eq:taylor}
 \end{equation}
 where we have made $\hat{L}^h_{B_2}(\nabla B_1)$ the higher order terms.

We use $ \eta \nabla \hat{L}(B_2, w)\cdot \nabla B_1(w)$ to approximate Equation \ref{eq:realpen} so that the penalty can be written
 \begin{equation}
- \hat{L}^h_{B_2}(\nabla B_1) =  \hat{L}(B_2, w) - \hat{L}(B_2, w - \nabla B_1)   - \eta \nabla B_1 \cdot \nabla B_2
\label{eq:highorder}
\end{equation}

 In Section \ref{sec:pen} we discussed the penalty in the case of GD where the entire training data is shown in every round. However, as shown in previous work, selecting a minibatch introduces additional noise. The batches in SGD have different relationships to the parameter $w$.    We call the batch being used to perform the gradient update the updating batch. In Equation \ref{eq:taylor}, $B_1$ is the updating batch.  As discussed in Section \ref{sec:expbev}, we expect $|\hat{L}^h_{B_1}(\nabla B_1)| > |\hat{L}^h_{B_2}(\nabla B_1)|$, and $\nabla B_1 \cdot \nabla B_1 > \nabla B_1 \cdot \nabla B_2$. We will discuss this effect in more detail in Section \ref{sec:batch}.
 
We are also interested in the behavior of larger versus smaller capacity models. We would expect that since larger models generalize better than smaller ones, that even if $\nabla B_1 \cdot \nabla B_1$ is larger for larger models,  $|\hat{L}^h_{B_2}(\nabla B_1)|$ or $|\hat{L}^h_{B_1}(\nabla B_1) |$ is also larger for larger models providing a regularization effect. We will discuss this more in Section  \ref{sec:modelcomp}

  \section{Plotting the penalty term and dot product over different batches}
  
 \textbf{Notation} Generically, we will use $B_u$ to refer to the updating batch $B_r$ to refer to a recently updating batch and $B_a$ to a long ago updating batch. We will also use $\Delta L_a = \hat{L}(B_a,w) - \hat{L}(B_a,w - \nabla B_u)$, $\Delta L_r = \hat{L}(B_r, w) - \hat{L}(B_r, w- \nabla B_u)$ and $\Delta L_u = \hat{L}(B_u, w) - \hat{L}(B_u, w - \nabla B_u)$
  
  \textbf{Gist:} In this section, we will show that the updating batch 1) is able to claim a larger reward from Equation \ref{eq:realpen} than other batches, but 2) also experiences a larger penalty for doing so. We conclude that the penalty penalizes the updating batch, which seems most at risk for being overfitted in a particular round.
 
  \label{sec:batch}
  
    \begin{figure*}[ht]
\centering
\includegraphics[scale=0.27]{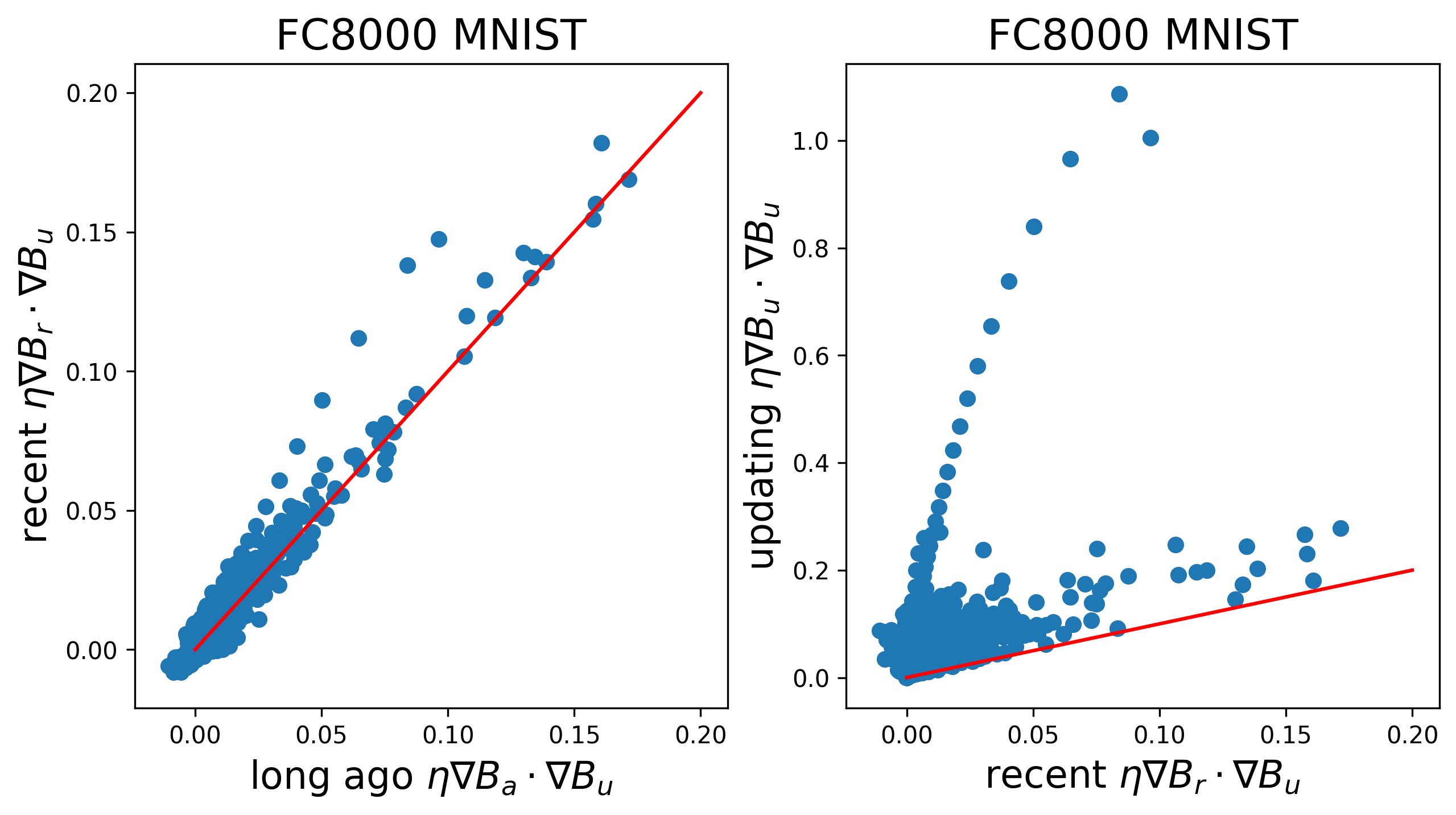}
\includegraphics[scale=0.27]{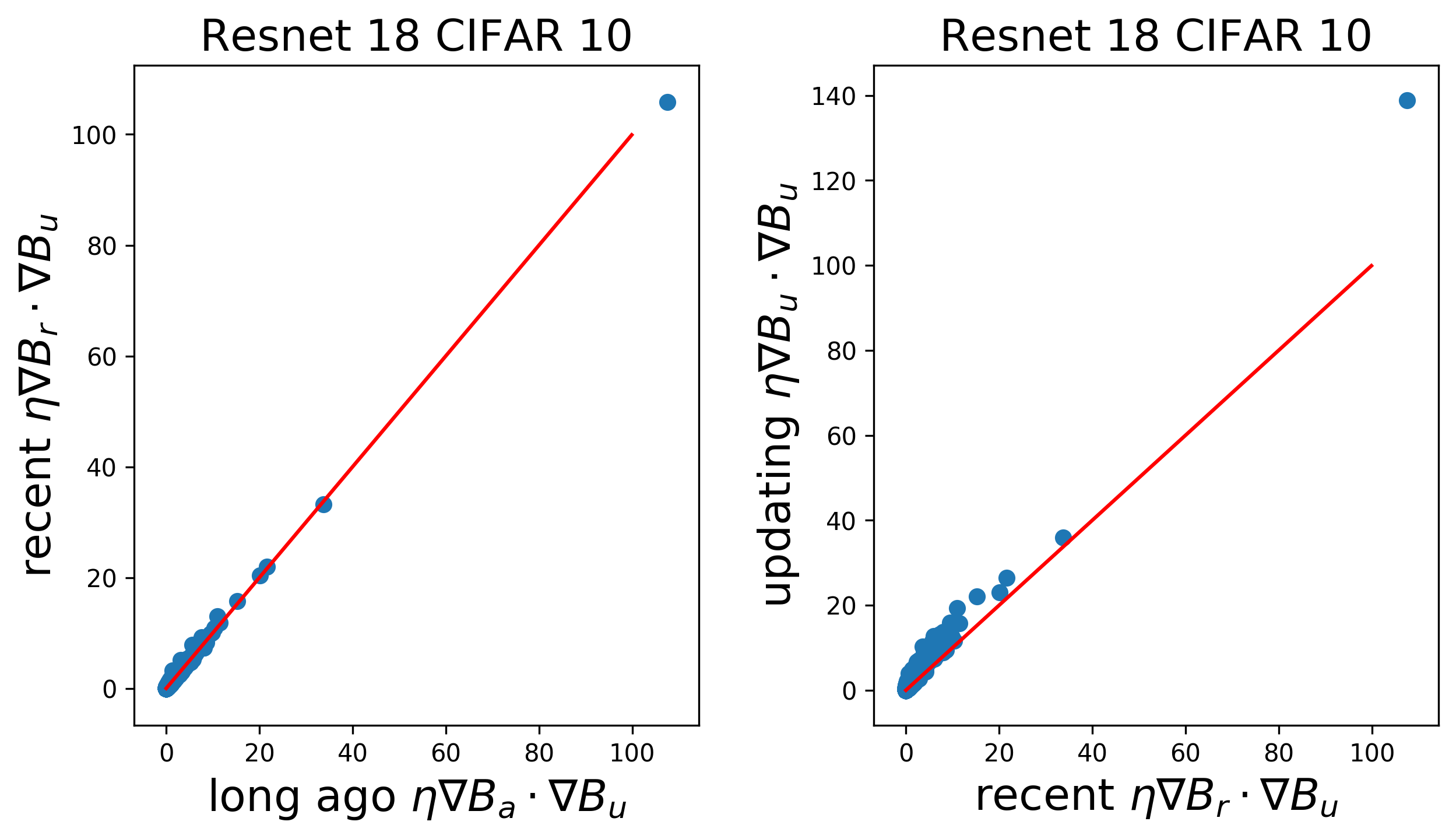}
\caption{ Red line depicts y=x. Updating versus long ago updating batches and recently updating batches. \textbf{Column 1}: $\eta \nabla B_r \cdot \nabla B_u$ versus $\eta \nabla B_a \cdot \nabla B_u$ for FC8000. \textbf{Column 2}:  $\eta \nabla B_u \cdot \nabla B_u$ versus $\eta \nabla B_r \cdot \nabla B_u$  for FC8000. \textbf{Column 3}:   $\eta \nabla B_r \cdot \nabla B_u$ versus $\eta \nabla B_a \cdot \nabla B_u$ for Resnet 18 on CIFAR 10. \textbf{Column 4}:   $\eta \nabla B_u \cdot \nabla B_u$ versus $\eta \nabla B_r \cdot \nabla B_u$ for Resnet 18 on CIFAR 10. We find that dot products for the updating batch on itself tend to be higher than dot products between the updating batch and a recently or long ago updating batch.}
\label{fig:longupcomp}
\end{figure*}

    Figure \ref{fig:longupcomp} shows  $\eta \nabla B_r \cdot \nabla B_u$ vs $\eta \nabla B_a \cdot \nabla B_u$ in Column 1 and 3 and  and  $\eta \nabla B_u \cdot B_u$  versus $\eta \nabla B_r \cdot \nabla B_u$  in Columns 2 and 4 for a fully connected two layer 8000 node network on MNIST in the left two columns and a Resnet 18 on CIFAR 10 in the right two columns. Consistently with what we would expect, we find  $\eta \nabla B_u \cdot \nabla {B_u}$ , is higher than   $\eta \nabla B_a \cdot \nabla {B_u}$ and $\eta \nabla B_r \cdot \nabla B_u$.
  
  Next, we examine the penalty term. Figure \ref{fig:nondot} depicts  in Row 1, from left to right,  $-\hat{L}^h_{B_r}(\nabla B_u)$ versus $-\hat{L}^h_{B_a}(\nabla B_u)$for FC 8000 on MNIST followed by $-\hat{L}^h_{B_u}(\nabla B_u)$ versus $-\hat{L}^h_{B_r}(\nabla B_u)$ for FC8000 on MNIST  followed by $-\hat{L}^h_{B_r}(\nabla B_u)$ versus $-\hat{L}^h_{B_a}(\nabla B_u)$ for Alexnet on CIFAR 10 followed by $-\hat{L}^h_{B_u}(\nabla B_u)$ versus $-\hat{L}^h_{B_r}(\nabla B_u)$ for Alexnet on CIFAR 10. In  Row 2  it depicts $-\hat{L}^h_{B_r}(\nabla B_u)$ versus $-\hat{L}^h_{B_a}(\nabla B_u)$ for Resnet 18 on CIFAR 10 followed by  $-\hat{L}^h_{B_u}(\nabla B_u)$ versus $-\hat{L}^h_{B_r}(\nabla B_u)$ for Resnet 18 on CIFAR 10.
For all cases, as expected in Section \ref{sec:expbev}.we see that $| \hat{L}^h_{ B_u}( \nabla B_u)| \geq |\hat{L}^h_{ \nabla B_r}(\nabla B_u)|\geq |\hat{L}^h_{B_a}(\nabla B_u)|$  in all cases.

From this, we conclude that the updating batch is able to make more progress on itself because of its success in the first order, but it also incurs a large penalty because the weights do not work as well together as they do individually .

Therefore, even if the updating batch can cause the weights to overfit on its data in a first order sense, it has an increased higher order penalty that penalizes weight changes that may not generalize well.

\section{Comparing different model dot product and reduction in loss}

\label{sec:modelcomp}

  \begin{figure*}[ht]
\centering
\includegraphics[width=.87\textwidth, height=6.0cm]{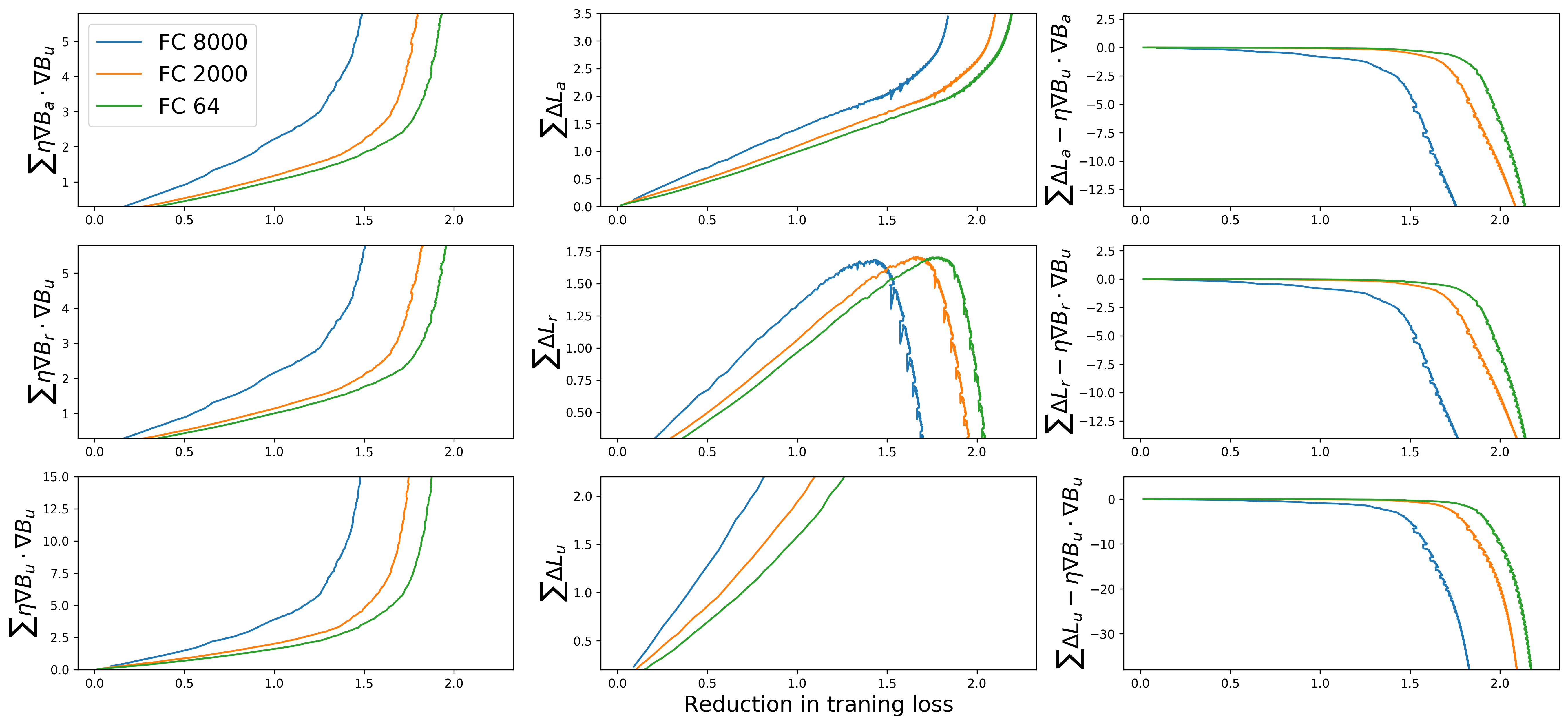}
\caption{Three two layer fully connected networks with 8000(blue), 2000(orange) and 64(green) units on MNIST. \textbf{Row 1}(left to right): $\sum \eta \nabla B_a \cdot \nabla B_u$ followed by $\sum \Delta L_a$ followed by $\sum \Delta L_a - \eta \nabla B_a \cdot \nabla B_u$ . \textbf{Row 2}(left to right):  $\sum \eta \nabla B_r \cdot \nabla B_u$ followed by $\sum \Delta L_r$ followed by $\sum \Delta L_r - \eta \nabla B_r \cdot \nabla B_u$.  \textbf{Row 3}(left to right):  $\sum \eta \nabla B_u \cdot \nabla B_u$ followed by $\sum \Delta L_u$ followed by $\sum \Delta L_u - \eta \nabla B_u \cdot \nabla B_u$ We find that larger models have more negative penalties, even though they also have higher dot products. }
\label{fig:mncomp}
\end{figure*}

\textbf{Gist:} We wish to compare how larger versus smaller capacity models behave in terms of the penalty. We expect that larger models are more heavily regularized, even if they are able to claim a larger reward from Equation \ref{eq:realpen}.

 We plot the reduction in loss so far on the x axis in order to compare the models at similar stages in training. We run an experiment on MNIST using $\eta=0.1$ for a fully connected two layer network with $64$(green), $2000$(orange), and $8000$(blue) nodes respectively.
The results are shown in Figure \ref{fig:mncomp}
 
 Figure \ref{fig:mncomp} shows the results of plotting in Row 1, $\sum \eta \nabla B_a \cdot \nabla B_u$ followed by $\sum \Delta L_a$ followed by $\sum \Delta L_a - \eta \nabla B_a \cdot \nabla B_u$. In Row 2, $\sum \eta \nabla B_r \cdot \nabla B_u$ followed by $\sum \Delta L_r$ followed by $\sum \Delta L_r - \eta \nabla B_r \cdot \nabla B_u$. And in Row 3  $\sum \eta \nabla B_u \cdot \nabla B_u$ followed by $\sum \Delta L_u$ followed by $\sum \Delta L_u - \eta \nabla B_u \cdot \nabla B_u$.
 
 First we notice that $\eta \nabla B_u \cdot \nabla B_u$, and $\eta \nabla B_u \cdot \nabla B_a$, and $\eta \nabla B_u \cdot \nabla B_r$ are higher for larger models as can be seen in Column 1 (we will actually find that for more complex datasets they are higher for a majority of training, but become lower at a later point in training).   We interpret this to mean that larger models can fit more in a first order sense, i.e. they are more able to find weights that would reduce the loss if implemented individually, and would therefore be able to increase the reward given by Equation \ref{eq:realpen}. However, we notice that for the recently updating batches, larger models experience a higher dot product, but experience a larger \emph{increase} in loss, and therefore a higher magnitude  penalty $|\hat{L}^h_{B_r}(\nabla B_u)|$. Larger models also experience a larger penalty over training as can be seen in Column 3 (again we will find that for more complex datasets this stops being true late in training.)
 
   \begin{figure*}[ht]
\centering
\includegraphics[scale=0.27]{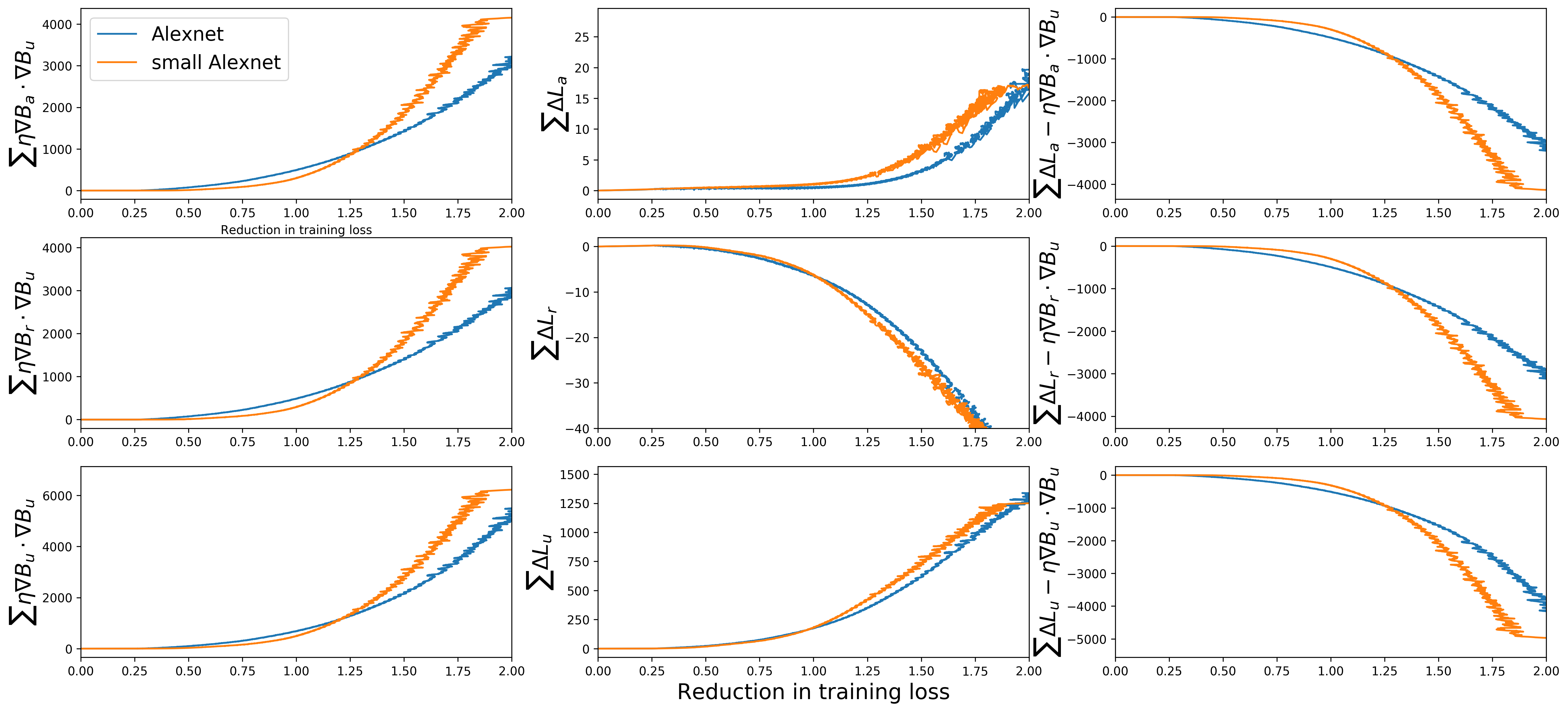}
\caption{Alexnet and small Alexnet on CIFAR 10. \textbf{Row 1}(left to right): $\sum \eta \nabla B_a \cdot \nabla B_u$ followed by $\sum \Delta L_a$ followed by $\sum \Delta L_a - \eta \nabla B_a \cdot \nabla B_u$ . \textbf{Row 2}(left to right):  $\sum \eta \nabla B_r \cdot \nabla B_u$ followed by $\sum \Delta L_r$ followed by $\sum \Delta L_r - \eta \nabla B_r \cdot \nabla B_u$.  \textbf{Row 3}(left to right):  $\sum \eta \nabla B_u \cdot \nabla B_u$ followed by $\sum \Delta L_u$ followed by $\sum \Delta L_u - \eta \nabla B_u \cdot \nabla B_u$ We find that larger models have more negative penalties for the bulk of training, even though they also have higher dot products.}
\label{fig:alexcomp}
\end{figure*}

We also show the analogue of Figure \ref{fig:mncomp} for Alexnet (blue)  and Alexnet with only one fully connected layer of size 256 (orange) on CIFAR 10 in Figure \ref{fig:alexcomp}. We use vanilla SGD and no momentum, batch normalization, or dropout. We use a constant learning rate of .01. We see in Figure \ref{fig:alexcomp} that the  larger model (Alexnet) has a larger magnitude penalty and dot product, until the models reach a training loss of about 1.0. We notice that this tracks the time the models begins overfitting in the test loss (see Appendix Figure \ref{fig:tloss})
 
   \begin{figure*}[ht]
\centering
\includegraphics[scale=0.3]{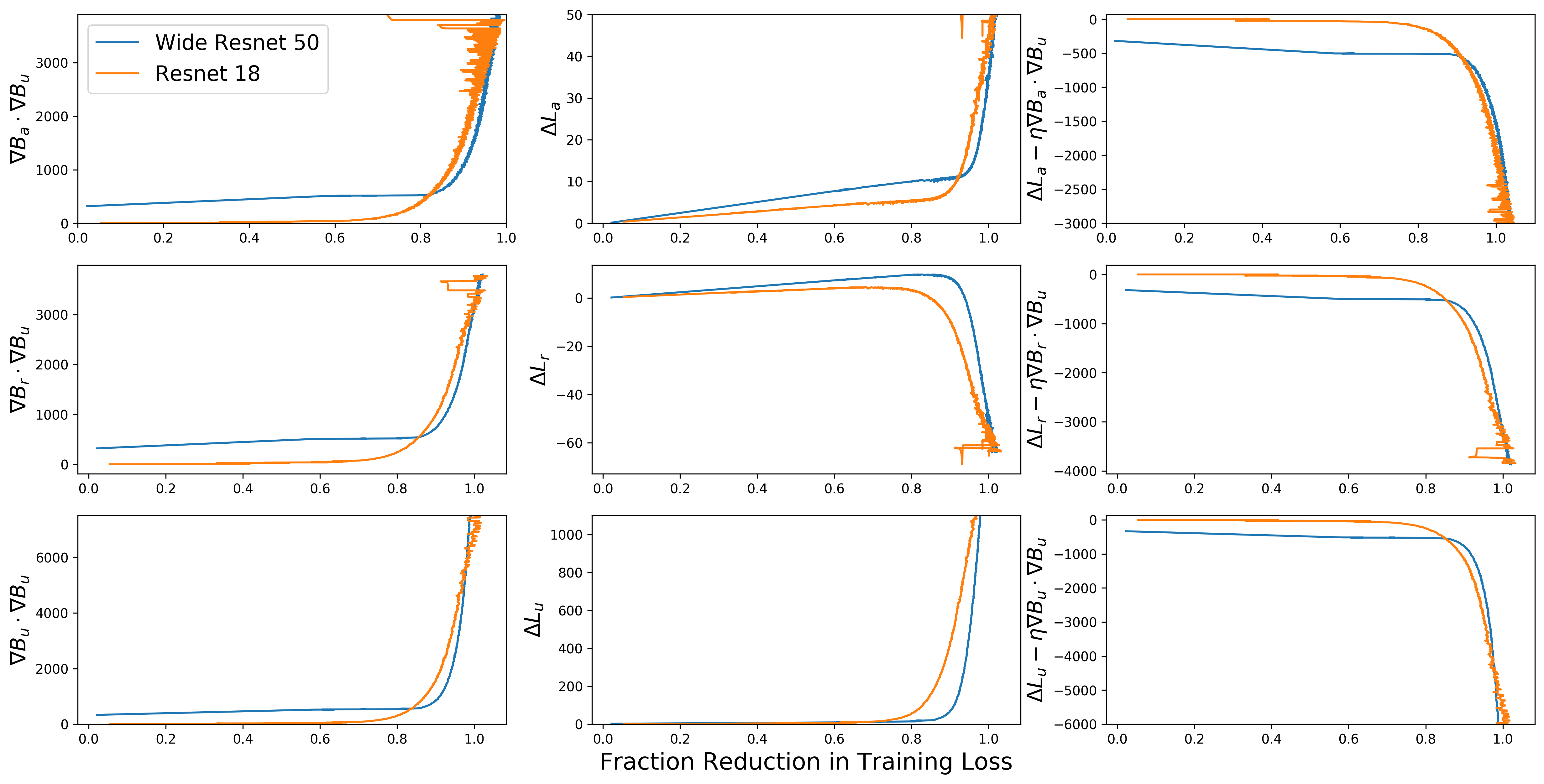}
\caption{ Resnet 18 and Wide Resnet 50 on CIFAR 10 \textbf{Row 1}(left to right): $\sum \eta \nabla B_a \cdot \nabla B_u$ followed by $\sum \Delta L_a$ followed by $\sum \Delta L_a - \eta \nabla B_a \cdot \nabla B_u$ . \textbf{Row 2}(left to right):  $\sum \eta \nabla B_r \cdot \nabla B_u$ followed by $\sum \Delta L_r$ followed by $\sum \Delta L_r - \eta \nabla B_r \cdot \nabla B_u$.  \textbf{Row 3}(left to right):  $\sum \eta \nabla B_u \cdot \nabla B_u$ followed by $\sum \Delta L_u$ followed by $\sum \Delta L_u - \eta \nabla B_u \cdot \nabla B_u$ We find that larger models have more negative penalties for the bulk of training, even though they also have higher dot products.}
\label{fig:rescomp}
\end{figure*}

 We also show the analogue of Figure \ref{fig:mncomp} for Resnet 18 (orange) versus a wide Resnet 50 (blue) on CIFAR 10 in Figure \ref{fig:rescomp}. We train a Resnet 18 model and a wide Resnet 50 model with width factor 2. We use a learning rate of .01 and a batch size of 150. 
We plot on the x-axis the fraction reduction in loss. We observe similar results to the Alexnet case.

\section{Conclusion}

We identify a property of using gradient based optimizers, namely that they update all the scalar weights at the same time instead of updating them individually. We find that this introduces uncertainty into the optimization, as each scalar weight knows the values of the other weights at $\mathbf{w_t}$ but is then evaluated at $\mathbf{w_{t+1}}$. We relate this phenomenon to the Taylor series. We find that penalties are most pronounced for batches that are currently being used to update. We find that penalties are higher for larger models.   Examining a broader array of datasets and architectures, and studying how this phenomenon interacts with other regularizers such as batch normalization and skip connections is an interesting investigation we leave to future work.

 \newpage

\bibliography{neurips_2019}
\bibliographystyle{unsrtnat}

\appendix
\clearpage

\section{Appendix}

\subsection{Small Alexnet architecture}

\begin{figure*}[ht]
\centering
\includegraphics[scale=0.6]{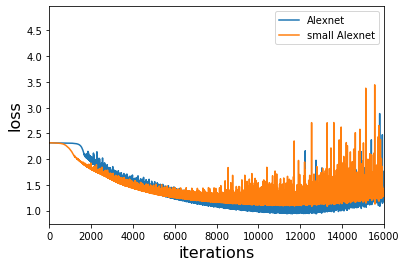}
\caption{ Test loss for Alexnet and small Alexnet}
\label{fig:tloss}
\end{figure*}

Our small Alexnet model retains the original Alexnet convolutional layers, but replaces the fully connected layers by a single one with 256 nodes.
We use a batch size of 150 and a constant learning rate of 0.01 for CIFAR 10 experiments. We depict the test loss for Alexnet (blue) and small Alexnet (orange) in Figure \ref{fig:tloss}. By comparing with Figure \ref{fig:alexcomp} we see that the crossover point, where Alexnet starts to have smaller dot product than small Alexnet, approximately tracks the point where Alexnet begins overfitting.

\newpage

\end{document}